\documentclass[11pt,a4paper]{article}
\usepackage[T1]{fontenc}
\usepackage[utf8]{inputenc}
\usepackage{lmodern}
\usepackage[margin=25mm]{geometry}
\usepackage{microtype}
\usepackage{amsmath,amssymb,bm}
\usepackage{array,booktabs,tabularx}
\usepackage{graphicx}
\usepackage[section]{placeins}
\usepackage{needspace}
\usepackage{cite}
\usepackage{xurl}
\usepackage[hidelinks]{hyperref}
\usepackage[font=small,labelfont=bf]{caption}
\newcommand{\method}{MuJoCable}
\newcommand{\degree}{^{\circ}}
\hypersetup{
  pdftitle={MuJoCable: Reduced-Order Surface-Routed Cable Transmission for Tendon-Driven Robots},
  pdfauthor={Yi Zhang, Qi Shao, Yicong Lin, Muyuan Ma, Tao Sun, Yue Xie},
  pdfsubject={Robotics; cable transmission simulation},
  pdfkeywords={MuJoCo, tendon-driven robots, cable routing, friction, simulation}
}
\title{\LARGE\bfseries MuJoCable: Reduced-Order Surface-Routed Cable Transmission for Tendon-Driven Robots}
\author{Yi Zhang, Qi Shao, Yicong Lin, Muyuan Ma,\\
Tao Sun\thanks{Corresponding author.}, and Yue Xie}
\date{}
\begin{document}
\maketitle
\raggedbottom

\begin{abstract}
Tendon transmissions reduce distal inertia and add compliance, yet routing, slack, and friction govern motion and force transfer. Mainstream rigid-body robotics simulators such as MuJoCo do not jointly resolve moving noncircular contact, unilateral tension, and segment friction. We present \method, which adds a reduced-order, configuration-dependent cable transmission to MuJoCo. Its routing algorithm jointly optimizes an ordered path across moving analytic and mesh surfaces. A unilateral axial law, directional Capstan propagation, and nodal virtual work map this path to segment tensions and body forces. The warm-started engine plugin applies these forces during simulation and exposes route and load states for design. Pulley benchmarks recover analytical transmission relations with a Capstan-ratio error below 0.5\%. On the underactuated 18-joint SpiRobs, \method\ reveals friction-driven load growth and proximal redistribution of joint rotation that the native tendon does not represent. Hardware tests on SpiRobs and a tendon-route-coupled finger reproduce observed motion sequences. By making physical threading executable, \method\ brings transmission sources of the simulation-to-reality gap into route, cable, and actuator design before fabrication.
\end{abstract}

\section{Introduction}
\label{sec:introduction}

\begin{figure}[t]
  \centering
  \includegraphics[width=\linewidth]{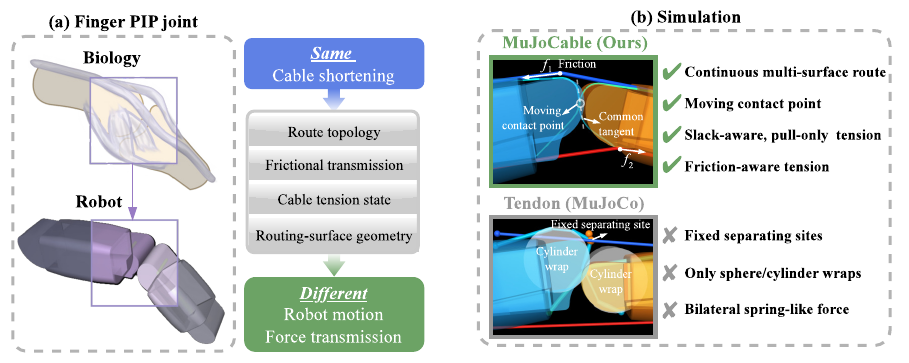}
  \caption{\textbf{Why cable transmission must be modeled by MuJoCable.} (a) Route topology, frictional transmission, cable tension state, and routing-surface geometry can map the same cable shortening to different robot motion and force transmission. (b) \method\ resolves a continuous path over moving surfaces with unilateral, friction-aware tension.}
  \label{fig:teaser}
\end{figure}

Tendon actuation places motors away from distal links, which reduces moving inertia and allows compact, compliant mechanisms. Dexterous hands, continuum robots, and rolling-contact robots all use this architecture~\cite{toshimitsu2023rolling,decker2026rolling,wang2025spirobs}. In each case, the cable forms the mechanical transmission between the actuator and the structure.

Motor retraction changes cable length and tension, and robot motion changes the cable path in return. As contact points migrate, moment arms vary. Holes and surface wraps also create unequal segment tensions~\cite{wang2013tendon,do2014friction}. Underactuation distributes both effects across several joints, so the same motor retraction can produce different joint torques and robot shapes. Cable routing is therefore part of the mechanism and its control interface.

Figure~\ref{fig:teaser} summarizes this coupling. Route topology, routing-surface geometry, frictional transmission, and cable tension state together determine how an actuator command becomes robot motion and force.

A useful robot simulator must resolve the transmission states behind this coupling. Many reduced-order models prescribe a route or encode an equivalent joint law. They are efficient, but contact migration cannot emerge from adjacent noncircular surfaces. MuJoCo spatial tendons compute minimum-length paths through body-fixed sites and analytic wraps~\cite{todorov2012mujoco}. The available wrapping obstacles are spheres and cylinders, and multiple obstacles require separating sites. Native tendon stiffness follows a Hookean law about springlength. With a single rest length, the force reverses across that length and can therefore generate both pulling and pushing generalized forces instead of enforcing a unilateral cable state. These choices do not expose continuous contact migration over adjacent noncircular surfaces or guide-wise tension states. A physical cable can also go slack and develop unequal segment tensions. Distributed rod models instead resolve cable mass, bending, torsion, and free shape~\cite{bergou2008der,naughton2021elastica,choi2024dismech,chen2025der}. Those states are appropriate for sag, buckling, knotting, and wave motion. Short, taut robot transmissions need moving surface contact and segment tension without the state dimension of a free cable.

We present \method, a reduced-order surface-routed cable transmission model implemented as a MuJoCo engine plugin. Physical threading and moving surfaces define a configuration-dependent route whose length drives a unilateral axial law. Directional Capstan relations propagate tension along the route, and virtual work maps the resulting nodal forces to rigid-body dynamics. The solved route also provides sensing, diagnostics, and visualization.

The evaluation follows the same causal sequence. Pulley systems compare the transmission mechanics with closed-form references. Robot simulations then test moving noncircular surfaces and guide friction. Finally, simulation-to-hardware comparisons evaluate a printed SpiRobs prototype and a tendon-coupled underactuated finger. For these two cable-driven underactuated robots, cable routing and friction directly shape the motion distribution. The evidence supports the following contributions.

\noindent Our contributions are:
\begin{enumerate}
  \item \textbf{Configuration-dependent surface routing algorithm.} A joint contact optimization follows an ordered cable route across adjacent moving analytic and mesh surfaces. Cable stroke and moment arms emerge from the geometry of noncircular rolling contacts.
  \item \textbf{Frictional cable-to-body dynamics.} A pull-only axial law, directional Capstan propagation, and nodal virtual-work loading resolve slack, unequal segment tensions, actuator demand, and joint-level motion redistribution.
  \item \textbf{MuJoCo engine-plugin implementation.} An MJCF-configurable plugin warm-starts the route solve and reports route validity, take-up, slack, segment tension, and actuator load. Pulley tests, two robot comparisons, and runtime measurements evaluate both the mechanics and the implementation.
\end{enumerate}

Section~\ref{sec:related-work} reviews cable representations and frictional transmission models. Section~\ref{sec:method} presents surface routing, unilateral tension, friction propagation, and rigid-body coupling. Section~\ref{sec:experiments} evaluates pulley mechanics, robot-level effects, and physical motion.

\section{Related Work}
\label{sec:related-work}
\subsection{Cable representations in robot simulation}
Tendon models support underactuated hands~\cite{catalano2014softhand}, tendon-driven manipulators~\cite{choi2020ambidex}, continuum robots~\cite{webster2010continuum,shentu2026rigidbody}, and rolling-contact mechanisms~\cite{toshimitsu2023rolling,decker2026rolling}. Open parametric hands also coordinate morphology and actuation through tendon-length and pulley mappings~\cite{gilday2024embodied}.

General-purpose rigid-body robotics simulators provide contact, actuation, and policy-development workflows~\cite{todorov2012mujoco,lee2018dart,mittal2023orbit}. In MuJoCo, a spatial tendon is a scalar-length element built from body-fixed sites and analytic wrap objects~\cite{todorov2012mujoco}. The native formulation maps this path to one transmission force. It does not expose continuous contact migration over adjacent noncircular surfaces or segment-wise frictional tensions.

OpenSim-style methods compute massless muscle paths from fixed, conditional, or moving points and analytic wrap objects~\cite{delp2007opensim,scholz2016wrapping}. Specialized methods extend this construction to arbitrary surfaces~\cite{lloyd2021wrapping}. These paths provide muscle length and moment arms, but not direction-dependent losses and unequal tensions across robot guides. MyoSim maps musculoskeletal kinematics and force parameters into MuJoCo for contact-rich simulation, but does not add a moving noncircular cable-contact model~\cite{wang2022myosim}. EquiMus uses energy-equivalent discretization to retain actuator mass and elastic energy in musculoskeletal robot dynamics~\cite{zhu2025equimus}. Distributed rods instead retain mass, stretch, bend, twist, contact, and free shape~\cite{bergou2008der,naughton2021elastica,choi2024dismech,chen2025der}. They resolve free-cable deformation by introducing distributed states. Short, taut robot transmissions require moving surface contact, unilateral tension, and segment friction without those free-cable states.

\begin{table}[t]
\caption{Cable representations for tendon-driven robots.}
\label{tab:landscape}
\centering
\scriptsize
\setlength{\tabcolsep}{3.2pt}
\renewcommand{\arraystretch}{1.15}
\begin{tabularx}{\linewidth}{@{}>{\hsize=1.55\hsize\linewidth=\hsize\raggedright\arraybackslash}X >{\hsize=1.5\hsize\linewidth=\hsize\raggedright\arraybackslash}X *{5}{>{\hsize=0.79\hsize\linewidth=\hsize\centering\arraybackslash}X}@{}}
\toprule
Approach & Cable representation &
Irregular moving surface routing &
Unilateral slack &
Unequal segment tensions &
Distributed mass and bending &
Rigid-body coupling \\
\midrule
Equivalent joint or kinematic model~\cite{webster2010continuum,toshimitsu2023rolling,gilday2024embodied}
& Joint-level length, torque, or transmission relation
& -- & dependent & dependent & -- & $\checkmark$ \\

MuJoCo spatial tendon~\cite{todorov2012mujoco}
& Body-fixed sites, sphere/cylinder wraps, and pulley branches
& -- & configured & -- & -- & $\checkmark$ \\

Musculoskeletal wrapping~\cite{delp2007opensim,scholz2016wrapping,lloyd2021wrapping,wang2022myosim}
& Massless path with via points and wrapping surfaces
& extended & muscle model & -- & -- & $\checkmark$ \\

DER or Cosserat rod~\cite{bergou2008der,naughton2021elastica,choi2024dismech,chen2025der}
& Distributed rod states with contact
& contact model & emergent & contact model & $\checkmark$ & dependent \\

\textbf{\method\ (ours)}
& \textbf{Massless path over moving contact surfaces}
& $\checkmark$ & $\checkmark$ & $\checkmark$ & -- & $\checkmark$ \\
\bottomrule
\end{tabularx}
\end{table}

\subsection{Frictional transmission models}
Friction in sheaths, holes, eyelets, pulleys, and surface wraps changes tension from one cable segment to the next. Analytical models describe curved sheaths, elasticity, backlash, presliding, and hysteresis~\cite{wang2013tendon,do2014friction,jung2008capstan,daemi2023friction}. Complementarity models resolve static and sliding contact at tendon disks~\cite{shen2026friction}, and learned models infer distal force or residual dynamics from measurements~\cite{li2019deep,choi2020ambidex,liu2024pinn}. These methods address friction after choosing a sheath centerline, disk sequence, or learned input representation. They do not by themselves generate the moving multi-surface route that determines wrap angles and body loads. Compact transmissions often operate in sustained sliding. In that regime, the Capstan relation links wrap angle, friction, direction, and segment tension directly. It therefore makes guide friction an explicit design variable in simulation.

\section{Method}
\label{sec:method}

\subsection{Scope, assumptions, and target problem}

\method\ targets short, tension-dominated transmissions in rigid-body robots. Each cable is a massless, axially compliant line with a quasi-static route over known rigid surfaces~\cite{delp2007opensim,scholz2016wrapping,lloyd2021wrapping}. It carries tension only and may go slack. Contact can move within a selected route class. The friction law represents sustained sliding with a supplied or regularized direction and no presliding memory. MuJoCo handles body contact, constraints, and time integration.

At every rigid-body step, the method solves three coupled quantities. These are a valid cable path over the moving surfaces, the unilateral friction-dependent tension in each path segment, and the loads applied to the connected bodies. This is a reduced-order transmission problem. It retains surface routing and force transfer without adding distributed cable degrees of freedom. The model applies when routing and tension transmission dominate free-cable sag, bending, and wave motion.

The algorithmic task is to update contact points jointly across adjacent moving analytic and mesh surfaces. The engineering task is to execute that solve inside MuJoCo, apply pull-only segment loads to the owning bodies, and expose the transmission states through the robot model.

\subsection{Cable representation and problem formulation}

\begin{figure}[htbp]
  \centering
  \includegraphics[width=0.78\linewidth]{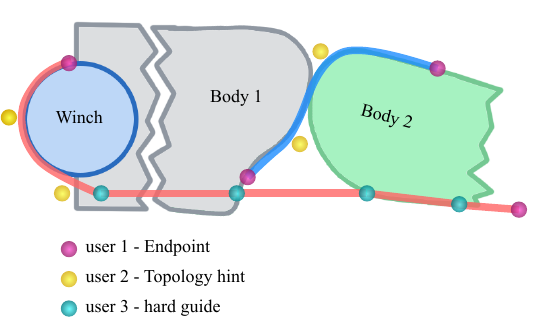}
  \caption{\textbf{Cable-site roles.} $user=2$ sites initialize surface routing, $user=3$ sites impose hard guides, and $user=1$ sites define cable endpoints.}
  \label{fig:cable-site-roles}
\end{figure}

A cable specification $\mathcal C$ contains endpoints, physical hard guides, routing surfaces, free-length and axial parameters, friction coefficients, and tension limits. An ordered topology seed $\mathcal H$ initializes its route class. Figure~\ref{fig:cable-site-roles} shows how these elements enter an MJCF model. A $user=1$ site defines an endpoint. A $user=3$ site imposes a physical hard guide. A $user=2$ site supplies the surface order, wrapping side, and mesh corridor. Runtime contact replaces this topology hint, so it carries no cable force. This separation keeps physical constraints distinct from route initialization.

At step $k$, the configuration is $\bm q_k\in\mathbb R^{n_q}$, the tangent velocity is $\bm v_k\in\mathbb R^{n_v}$, and the control is $u_k$. The map $\chi$ converts the command into a positive shortening $c_k$. The step-level transmission problem is
\begin{equation}
\begin{aligned}
c_k&=\chi(u_k,\mathcal C),\\
(\mathcal C,\mathcal H,\bm q_k,\bm v_k,c_k)
&\longmapsto
\{\gamma_k^*,L_k,G_k,\dot L_k,\{T_{i,k}\},\bm Q_{c,k}\}.
\end{aligned}
\label{eq:overview}
\end{equation}
Here $\gamma_k^*$ is the solved route, $L_k$ is its length, $G_k$ is its length Jacobian, and $\dot L_k$ is its rate of change. The tension in segment $i$ is $T_{i,k}$, and $\bm Q_{c,k}$ is the resulting generalized load. A valid route is represented by nodes $\bm x_{i,k}$, segment tangents $\hat{\bm t}_{i,k}$, and wrap angles $\phi_{i,k}$. The point Jacobian $\bm J_{i,k}$ maps the nodal force $\bm f_{i,k}$ to generalized coordinates.

The constitutive states are the free length $L_{f,k}$, effective extension $e_k$, source tension $T_k$, and taut indicator $\mathrm{taut}_k=\mathbf 1[T_k>0]$. The axial law uses stiffness $\kappa$, damping $d$, transition width $a$, and tension limit $T_{\max}$. Its smooth ramp $\psi_a$ and damping gate $\eta_a$ activate only in tension, while $\operatorname{clip}_{[l,h]}$ limits a value to $[l,h]$. Each routed contact uses friction coefficient $\mu_i$ and propagation direction $\sigma_i$.

\begin{figure}[t]
  \centering
  \includegraphics[width=0.98\textwidth]{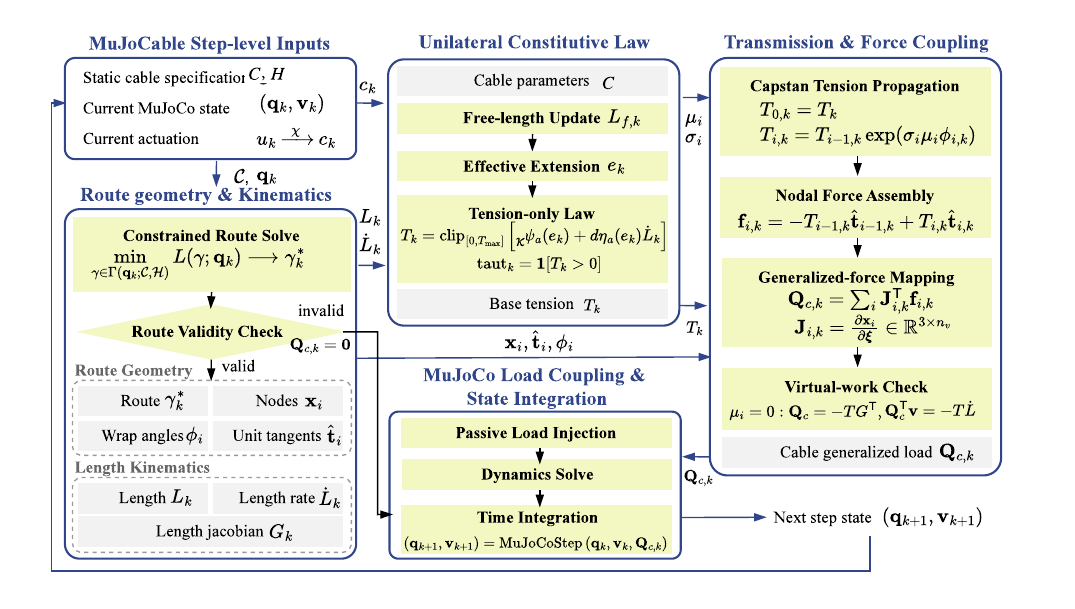}
  \caption{\textbf{MuJoCable framework.} Step-level cable and mechanism coupling. The static input is $(\mathcal C,\mathcal H)$, and subscript $k$ denotes the simulation step. Route-local labels omit $k$ where unambiguous.}
  \label{fig:pipeline}
\end{figure}

The set $\Gamma(\bm q,\mathcal C,\mathcal H)$ contains admissible routes, and $L(\gamma,\bm q)$ is the length of route $\gamma$. The coordinate $\bm\xi\in\mathbb R^{n_v}$ is local to the MuJoCo tangent space. The modules follow the same order as the outputs in~\eqref{eq:overview}. Route geometry drives the unilateral law, directional Capstan propagation gives the segment tensions, and nodal forces produce $\bm Q_{c,k}$. MuJoCo integrates this load and updates the moving surfaces, which closes the loop in Fig.~\ref{fig:pipeline}. We omit the step index below.

\subsection{Core transmission model}

\Needspace{5\baselineskip}
\noindent\textbf{Surface route and length kinematics.}
This module makes cable travel depend on the current surface geometry. Every curve in $\Gamma(\bm q,\mathcal C,\mathcal H)$ passes through each hard guide, satisfies surface nonpenetration, and belongs to $\mathcal H$.

\begin{equation}
\begin{aligned}
 \gamma^*(\bm q)
 &=\arg\min_{\gamma\in\Gamma(\bm q,\mathcal C,\mathcal H)}
 L(\gamma,\bm q),\\
 L(\bm q)&=L(\gamma^*(\bm q),\bm q)
 =\int_{\gamma^*}\mathrm ds .
\end{aligned}
 \label{eq:path}
\end{equation}
Here $\mathrm ds$ is the arc-length element. For the $N$ route segments, let $\delta\bm\xi\in\mathbb R^{n_v}$ be a MuJoCo tangent perturbation and let $\bm q\oplus\delta\bm\xi$ denote its application to the configuration.
\begin{equation}
\begin{aligned}
 L(\bm q)&=\sum_{i=0}^{N-1}\|\bm x_{i+1}-\bm x_i\|,\\
 G(\bm q)&=\left.\frac{\partial L(\bm q\oplus\delta\bm\xi)}
 {\partial\delta\bm\xi}\right|_{0},
 & \dot L&=G\bm v.
\end{aligned}
\label{eq:length}
\end{equation}
The gradient $G\in\mathbb R^{1\times n_v}$ remains defined when $n_q\ne n_v$. For a fixed contact topology, the route is locally stationary and the envelope theorem removes explicit derivatives of the internal contact parameters.

Cylinders use analytic cross-sectional tangencies. A surface segment becomes straight after unrolling it into $(r\theta,z)$, where $r$, $\theta$, and $z$ are the cylinder radius, circumferential angle, and axial coordinate. Closed meshes use an initialized triangle corridor whose edge crossings are optimized under nonpenetration at runtime. Nonconvex obstacle mode removes contacts with an adhesive normal force. Guided mode instead preserves a known support corridor. Adjacent moving surfaces share one optimization over their exit and entry points to avoid an artificial corner at the interface.

MuJoCo spatial tendons use body-fixed sites and analytic sphere/cylinder wraps, with separating sites between multiple obstacles~\cite{todorov2012mujoco}. In \method, topology sites initialize rather than fix the runtime path. Joint optimization across adjacent surfaces extends the path to moving noncircular contacts and mesh guides. The resulting cable stroke emerges from the solved geometry.

\smallskip
\Needspace{5\baselineskip}
\noindent\textbf{Unilateral cable response.}
This module separates taut transmission from slack cable. Let $L_0$, $c$, $p$, and $s$ denote the home length, contraction, pretension offset, and slack reserve. Let $\kappa$, $d$, $T_{\max}$, and $a$ denote axial stiffness, damping, the tension limit, and transition width. The free length $L_f$, effective extension $e$, and source tension $T$ are

\begin{equation}
\begin{aligned}
 L_f&=L_0-c-p, & e&=L-L_f-s,\\
 T&=\operatorname{clip}_{[0,T_{\max}]}
 \big(\kappa\psi_a(e)+d\eta_a(e)\dot L\big).
\end{aligned}
\label{eq:tension}
\end{equation}
The operator $\operatorname{clip}_{[l,h]}$ saturates its argument to $[l,h]$. The $C^1$ ramp $\psi_a$ is $0$ for $e\le0$, $e^2/(2a)$ for $0<e<a$, and $e-a/2$ for $e\ge a$. The gate $\eta_a(e)=\operatorname{clip}_{[0,1]}(e/a)$ scales damping. The taut indicator is $\mathrm{taut}=\mathbf{1}[T>0]$.

Native MuJoCo tendon stiffness follows a Hookean spring about springlength~\cite{todorov2012mujoco}. With a single rest length, its force reverses across that length and can pull or push instead of enforcing a unilateral cable state. Equation~\eqref{eq:tension} makes the taut--slack state explicit and combines slack reserve, pretension, damping activation, and tension limits with the solved surface route.

\smallskip
\Needspace{5\baselineskip}
\noindent\textbf{Directional segment friction.}
This module converts guide and wrap friction into unequal segment tensions. Set $T_0=T$. For wrap angle $\phi_i\ge0$, coefficient $\mu_i\ge0$, and supplied or regularized propagation direction $\sigma_i\in[-1,1]$, the sliding-limit relation gives~\cite{wang2013tendon,jung2008capstan}

\begin{equation}
 T_i=T_{i-1}\exp(\sigma_i\mu_i\phi_i).
 \label{eq:capstan}
\end{equation}
A native MuJoCo spatial tendon maps its scalar path to one transmitted force~\cite{todorov2012mujoco}. Equation~\eqref{eq:capstan} instead resolves the tension of every routed segment. Local friction therefore changes both actuator demand and the force delivered to each body.

\smallskip
\Needspace{5\baselineskip}
\noindent\textbf{Cable-to-body coupling.}
This module maps the routed segment tensions to the rigid-body dynamics. For segment direction $\hat{\bm t}_i=(\bm x_{i+1}-\bm x_i)/\|\bm x_{i+1}-\bm x_i\|$, an interior node and its point Jacobian satisfy

\begin{equation}
 \bm f_i=-T_{i-1}\hat{\bm t}_{i-1}+T_i\hat{\bm t}_i,
 \qquad
 \bm J_i=\left.\frac{\partial\bm x_i(\bm q\oplus\delta\bm\xi)}
 {\partial\delta\bm\xi}\right|_0 .
 \label{eq:node_force}
\end{equation}
The plugin applies $\bm f_i$ to the associated body through $mj\_applyFT$. For $\bm J_i\in\mathbb R^{3\times n_v}$, the equivalent generalized load is
\begin{equation}
 \bm Q_c=\sum_i\bm J_i^{\mathsf T}\bm f_i .
 \label{eq:generalized_force}
\end{equation}
For $\mu_i=0$, Eq.~\eqref{eq:capstan} gives $T_i=T$ and nodal assembly reduces to
\begin{equation}
 \bm Q_c=-T G^{\mathsf T},\qquad
 \bm Q_c^{\mathsf T}\bm v=-T\dot L,
 \label{eq:virtual_work}
\end{equation}
which provides force and power checks. If $c=c(\theta)$ and spool reaction is enabled, $\theta$ is the spool angle and virtual work gives its reaction torque $\tau_\theta=-T\,\partial c/\partial\theta$. A surface-mode seed tendon has zero actuator transmission, so the plugin is the single force path.

Equal tension recovers MuJoCo's scalar length-gradient mapping. When friction gives $T_i\ne T_{i-1}$, the nodal form also applies the unbalanced guide and wrap load to its owning body. This connects segment-level transmission loss to joint motion within the same rigid-body solve.

\smallskip
\Needspace{5\baselineskip}
\noindent\textbf{Route validity and diagnostics.}
Each step reports route status, slack, take-up, segment tension, actuator load, solver iterations, and residuals for tangency, surface error, and penetration. An admissible route remains active below its declared threshold. Penetration or topology failure sets $\bm Q_c=\bm0$ and retains the warm start. These states expose the physical route and transmission directly in the robot model.

\section{Experiments and Results}
\label{sec:experiments}

\subsection{Experiment Setup}

The evaluation has three levels, each tied to a different question. Theory-to-simulation tests whether the implemented transmission equations reproduce closed-form travel, force, tension-ratio, and torque relations in seven pulley systems. Simulation-to-simulation tests the representation enabled by the plugin. A rolling joint probes cable motion over moving noncircular surfaces, and the 18-joint logarithmic-spiral SpiRobs platform probes native-tendon equivalence and guide friction~\cite{wang2025spirobs}. Simulation-to-hardware then tests whether a robot built with the same geometry and cable route follows the simulated motion.

\subsection{Theory-to-Simulation Validation on Pulley Systems}

\begin{figure}[t]
  \centering
  \includegraphics[width=0.52\linewidth]{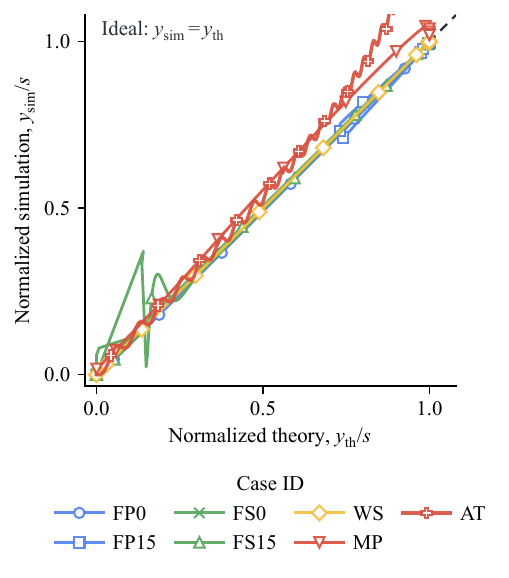}
  \caption{Normalized agreement between theory and simulation for pulley cases. IDs match Table~\ref{tab:result1_summary}.}
  \label{fig:pulley_validation}
\end{figure}

All seven pulley systems follow their analytical references for cable travel, force multiplication, Capstan transmission, pulley torque, and radius conversion. Figure~\ref{fig:pulley_validation} and Table~\ref{tab:result1_summary} summarize the errors.

The references use $\Delta l=n\Delta y$ and $F_{\mathrm{cable}}=nT$. Friction follows
\begin{equation*}
T_{i+1}/T_i=\exp[\mu\beta\tanh(v_{\mathrm{rel}}/v_s)].
\end{equation*} The remaining cases use free-sheave torque balance and $r_w\theta_w=-r_p\theta_p$.

Each case $i$ uses the analytical scale
\begin{equation}
s_i=\max_{t\in\Omega_i}\left|y_{\mathrm{theory},i}(t)\right|,
\end{equation}
followed by the normalized trajectory and error
\begin{equation}
\begin{aligned}
\widetilde y_i&=\frac{y_i}{s_i},\\
e_i(t)&=y_{\mathrm{sim},i}(t)-y_{\mathrm{theory},i}(t),\\
\mathrm{NRMSE}_i&=\frac{100}{s_i}
\left[\frac{1}{N_i}\sum_{t\in\Omega_i}e_i^2(t)\right]^{1/2}.
\end{aligned}
\end{equation}
where $\Omega_i$ contains the $N_i$ valid samples. Because the analytical response of FS0 is zero, its reference scale is the peak angular speed of FS15.

\begin{table}[htbp]
\caption{Pulley theory and simulation errors.}
\label{tab:result1_summary}
\centering
\setlength{\tabcolsep}{2pt}
\renewcommand{\arraystretch}{1.10}
\begin{tabular}{@{}>{\raggedright\arraybackslash}p{0.08\columnwidth}>{\raggedright\arraybackslash}p{0.21\columnwidth}>{\raggedright\arraybackslash}p{0.33\columnwidth}>{\raggedright\arraybackslash}p{0.28\columnwidth}@{}}
\toprule
ID & Case & Metric & Value \\
\midrule
FP0 & Fixed\newline \mbox{$\mu=0$} &
Travel RMSE\newline Tension RMSE &
0.161~mm (0.640\%)\newline $7.60{\times}10^{-3}$~N\\
FP15 & Fixed\newline \mbox{$\mu=0.15$} &
Peak ratio error\newline Tension RMSE &
$-6.10{\times}10^{-3}$ \newline 0.0165~N \\
FS0 & Free sheave\newline \mbox{$\mu=0$} &
Peak speed\newline Torque RMSE &
$3.74{\times}10^{-15}$~rad/s\newline $4.51{\times}10^{-17}$~N$\cdot$m \\
FS15 & Free sheave\newline \mbox{$\mu=0.15$} &
Peak speed error\newline Torque RMSE &
$-2.60{\times}10^{-4}$~rad/s\newline $4.09{\times}10^{-5}$~N$\cdot$m \\
WS & Winch-sheave &
Final lift error\newline Lift RMSE &
$-0.106$~mm\newline 0.0915~mm \\
MP & Moving pulley &
Final lift error\newline Lift RMSE &
$+0.185$~mm\newline 0.435~mm \\
AT & Atwood\newline pre-impact &
Contact-time error\newline Speed RMSE &
$-0.0358$~s\newline 0.0668~m/s \\
\bottomrule
\end{tabular}
\end{table}

\begin{figure}[t]
  \centering
  \includegraphics[width=0.70\linewidth]{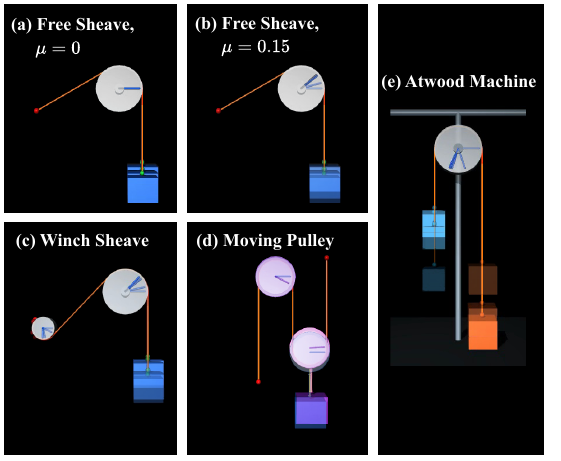}
  \caption{\textbf{Pulley motion overlays.} Friction drives FS15. WS, MP, and AT show radius transmission, lift, and pre-impact motion.}
  \label{fig:pulley_motion_overlays}
\end{figure}

The fixed-pulley cases verify cable travel and Capstan scaling. At $\mu=0$, the load-side tension matches the weight of the 0.2-kg payload.

The free-sheave cases show how friction creates motion (Fig.~\ref{fig:pulley_motion_overlays}(a),(b)). With $\mu=0$, equal segment tensions give $\tau=R(T_s-T_l)=0$, and the peak speed is $3.74\times10^{-15}$~rad/s. At $\mu=0.15$, the tension difference produces torque and the sheave reaches $0.605$~rad/s. The same segment law therefore accounts for both tension transmission and the resulting pulley motion.

The winch-sheave and moving-pulley cases follow their geometric ratios, and the Atwood case follows its closed-form motion before contact. Every route remains valid and unsaturated. The peak tangency residual is $7.2\times10^{-8}$, and the surface residual remains at floating-point scale.

\subsection{Simulation-to-Simulation Comparison}

Surface routing makes the cable path part of the robot dynamics. The rolling-joint case tests how moving surfaces change the mapping from cable stroke to joint motion. The SpiRobs comparison then examines the transmission states behind a similar global bend, followed by a friction sweep that measures changes in source load and shape.

\textbf{Rolling-contact joint.} The Faive PIP test compares two prescribed virtual hinges with a cable routed over two moving noncircular surfaces. The hinges prescribe $25\degree$ at each contact and produce $49.994\degree$ total rotation. With the same rolling geometry, a 10-mm shortening in \method\ produces $52.481\degree$. The route is valid for 99.92\% of the trajectory (Fig.~\ref{fig:robot_evaluation}(a)). The $2.487\degree$ difference reflects the cable-stroke mapping over the rolling surfaces rather than a prescribed joint relation.

\textbf{Native tendon comparison.} With the 18-joint SpiRobs morphology fixed~\cite{wang2025spirobs}, the native MuJoCo tendon and \method\ produce nearly the same steady bend. Yet \method\ resolves a peak cable tension nearly four times as large, while cable take-up is about 9\% higher (Fig.~\ref{fig:robot_evaluation}(c)). Similar global motion can therefore conceal different transmission states. Surface routing makes these states available for cable sizing, actuator selection, and route design.

\textbf{Guide-friction intervention.} The friction sweep keeps the geometry and controller fixed. Figure~\ref{fig:robot_evaluation} shows the resulting shapes and transmission states, while Table~\ref{tab:result2_friction} reports their values. At low friction, the overall bend remains nearly unchanged even as source tension begins to rise. Higher friction sharply increases the source load and progressively reduces bending. At $\mu=0.60$, peak tension is more than four times the frictionless value, while the robot retains about three quarters of its bending magnitude. Distal-joint motion also falls markedly as more rotation shifts toward the proximal joints. Guide friction thus changes both the source load and the distribution of robot motion.

\begin{table}[htbp]
\caption{SpiRobs guide-friction sweep.}
\label{tab:result2_friction}
\centering
\small
\setlength{\tabcolsep}{3.4pt}
\renewcommand{\arraystretch}{1.00}
\begin{tabular}{@{}rrrr@{}}
\toprule
$\mu$ & Steady bend (deg) & Peak tension (N) & Take-up (mm) \\
\midrule
0.00 & $-525.597$ & 1.856 & 87.590 \\
0.15 & $-522.970$ & 2.538 & 92.694 \\
0.30 & $-502.496$ & 6.553 & 98.907 \\
0.45 & $-438.513$ & 7.959 & 96.553 \\
0.60 & $-392.338$ & 7.951 & 90.041 \\
\bottomrule
\end{tabular}
\end{table}

A three-joint underactuated finger shows the same redistribution. Higher guide friction reduces final flexion from $269.3\degree$ to $214.3\degree$, and peak internal force falls from 0.395 to 0.183~N. Friction-aware routing therefore connects local transmission losses with robot-level shape before fabrication.

\Needspace{4\baselineskip}
\textbf{Runtime.} Seven 6-s headless runs on the matched 18-joint models give median step times of 34.25~$\mu$s for the native tendon and 41.96~$\mu$s for \method. The plugin adds 22.5\% to the step time and still runs at 11.9 times real time with a 0.5-ms step.\par

\subsection{Simulation-to-Hardware Validation}

\begin{figure}[p]
  \centering
  \includegraphics[width=0.88\textwidth]{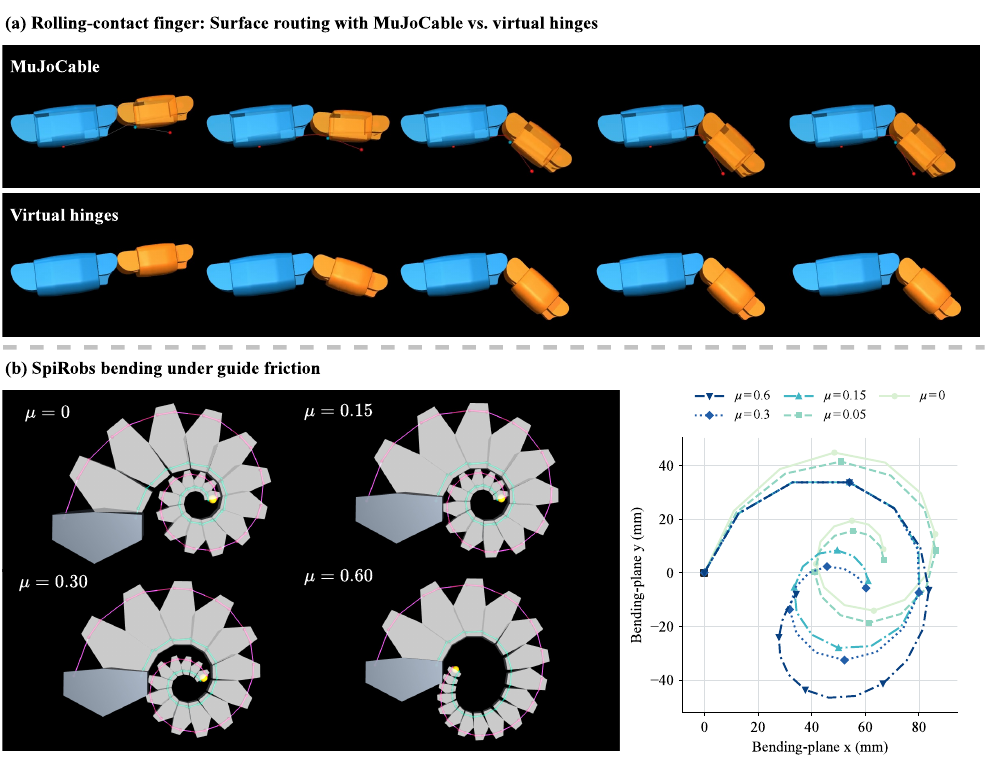}
  \vspace{2mm}

  \begin{minipage}[t]{0.48\textwidth}
    \vspace{0pt}
    \centering
    \includegraphics[width=0.86\linewidth]{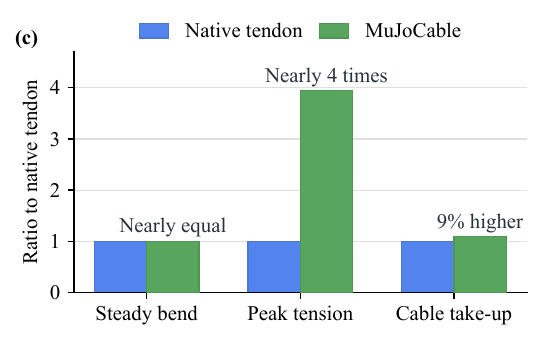}
    \vspace{2mm}
    \includegraphics[width=0.86\linewidth]{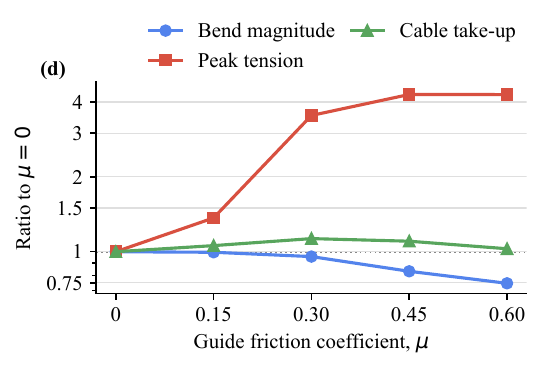}
  \end{minipage}
  \hfill
  \begin{minipage}[t]{0.48\textwidth}
    \vspace{0pt}
    \centering
    \includegraphics[width=\linewidth]{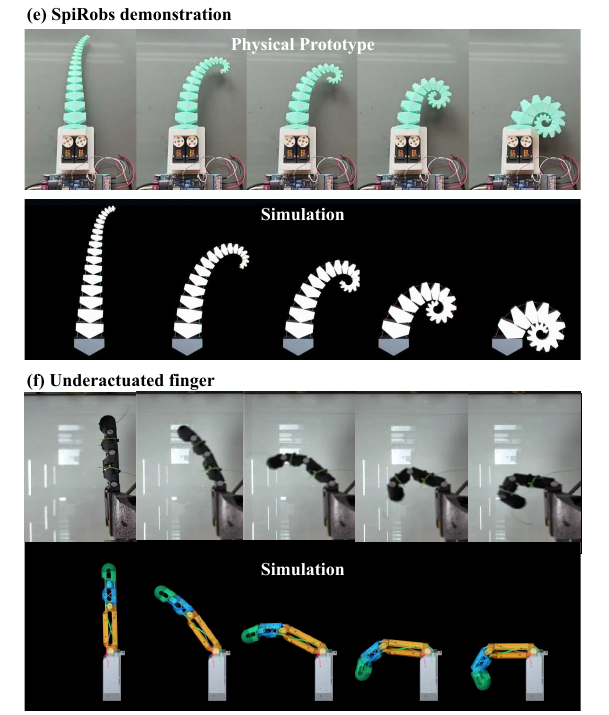}
  \end{minipage}
  \caption{\textbf{Robot-level evaluation.} (a) Rolling-contact motion. (b) SpiRobs shapes under guide friction. (c) Native-tendon transmission ratios. (d) Friction sweep. (e,f) SpiRobs and finger simulation-hardware sequences.}
  \label{fig:robot_evaluation}
\end{figure}

Figure~\ref{fig:robot_evaluation}(e,f) compares simulation and hardware. The SpiRobs platform~\cite{wang2025spirobs} uses the same geometry and cable route in both cases, and the overall bending sequences agree. The tendon-route-coupled underactuated finger~\cite{zhao2026prosthetic} also uses the same physical geometry and route.

\section{Discussion}
\label{sec:discussion}

These results show that cable routing is part of the mechanism in tendon-driven underactuated robots. One actuator drives several passive joints, so surface geometry, slack, and friction determine how cable stroke and tension are distributed. \method\ exposes this transmission layer. Designers can separate errors caused by geometry, take-up, or guide losses before changing the controller or fabricating another prototype.

The same states support component sizing. Material-dependent axial and friction parameters define the cable. Predicted take-up and source tension indicate actuator stroke and peak load, while segment tensions and tension limits support cable selection. Route validity and slack outputs allow threading, pretension, guide, and lubrication choices to be compared in one model. This moves transmission-induced reality gaps into early design decisions and shortens mechanical iteration.

MuJoCable helped refine the design of the underactuated finger. During prototyping, guide friction distorted motion and degraded control. Friction sweeps and route diagnostics identified critical guides and accumulated losses. Reducing friction at these locations and adding lubrication produced reliable tendon actuation. Simulation therefore directed a physical route revision rather than treating the discrepancy as a controller-only problem.

The route solver currently assumes a known route topology. It does not switch topology when a cable detaches, leaves a guide, or selects another wrap. Future work will add dynamic contact activation and topology transitions for derailment, re-engagement, and alternative threading-path design.

\section{Conclusion}
\label{sec:conclusion}

\method\ turns physical threading into an executable transmission model for general-purpose rigid-body robotics simulation. Its moving-surface route, unilateral cable law, and segment friction reproduce pulley mechanics and expose how cable geometry and guide losses redistribute robot load and motion. Hardware tests on the 18-joint SpiRobs platform and the tendon-route-coupled underactuated finger connect these transmission states to real underactuated motion. To our knowledge, \method\ is the first cable-transmission plugin for a general-purpose rigid-body robotics simulator to combine moving multi-surface routing, pulley transmission, unilateral tension, and guide friction in one robot model. It makes cable layout, pretension, material, and actuator choices accessible before fabrication, enabling physics-informed cable-route design. The same interface opens cable-driven robot design to route optimization and reinforcement-learning-based control.

\section*{Acknowledgment}
\textit{AI use disclosure.} OpenAI GPT-5.6 Sol, accessed through the Codex environment, assisted with software-architecture planning, automation of repeatability tests for the completed simulation models, and conversion of Type~3 fonts in Figs.~\ref{fig:teaser} and~\ref{fig:robot_evaluation} to submission-compatible vector graphics.

\bibliographystyle{IEEEtran}
\begingroup
\small
\sloppy
\bibliography{references}
\endgroup

\end{document}